\documentclass{bmvc2k}
\usepackage{multirow}

\title{Continual Vision-and-Language Navigation}

\addauthor{SeongJun Jeong}{jsj4968@snu.ac.kr}{1}
\addauthor{Gi-Cheon Kang}{chonkang@snu.ac.kr}{2}
\addauthor{Seongho Choi}{seongho.choi@sk.com}{3}
\addauthor{Joochan Kim}{joochan.k@kist.re.kr}{4}
\addauthor{Byoung-Tak Zhang}{btzhang@snu.ac.kr}{1,2*}

\addinstitution{
 Interdisciplinary Program in Artificial\\Intelligence, Seoul National University
}
\addinstitution{
 AI Institute for\\Seoul National University
}
\addinstitution{
 SK Telecom
}
\addinstitution{
 Korea Institute of Science\\and Technology
}

\runninghead{Jeong et al.}{Continual Vision-and-Language Navigation}

\begin{document}

\maketitle

\begin{abstract}
Developing Vision-and-Language Navigation (VLN) agents typically assumes a \textit{train-once-deploy-once} strategy, which is unrealistic as deployed agents continually encounter novel environments. To address this, we propose the Continual Vision-and-Language Navigation (CVLN) paradigm, where agents learn and adapt incrementally across multiple \textit{scene domains}. CVLN includes two setups: Initial-instruction based CVLN for instruction-following, and Dialogue-based CVLN for dialogue-guided navigation. We also introduce two simple yet effective baselines for sequential decision-making: Perplexity Replay (PerpR), which replays difficult episodes, and Episodic Self-Replay (ESR), which stores and revisits action logits during training. Experiments show that existing continual learning methods fall short for CVLN, while PerpR and ESR achieve better performance by efficiently utilizing replay memory.
\end{abstract}

\section{Introduction}
\label{sec:intro}
One of the desired attributes of robotic agents is their ability to learn and adapt to new environments without losing knowledge of previous ones. In practice, robots deployed in the real world continually encounter unseen environments and are required to perform a variety of tasks in them. Vision-and-Language Navigation (VLN) focuses on agents that navigate using natural language instructions and visual cues, promising for applications like robotic assistants. Despite recent advancements, most VLN research follows a \textit{train-once-deploy-once} strategy. A deployed VLN agent must promptly adapt whenever the environment changes throughout its lifetime. Finetuning on new environments might seem practical, but it can cause the agent's performance in previously learned contexts to deteriorate, a challenge known as catastrophic forgetting~\cite{mccloskey1989catastrophic}. To address this challenge, the agent needs the capability for Continual Learning (CL)~\cite{thrun2012learning}, which enables it to learn and adapt to new circumstances while retaining its existing knowledge.

\begin{figure*}[t]
\centering
\includegraphics[width=0.8\textwidth]{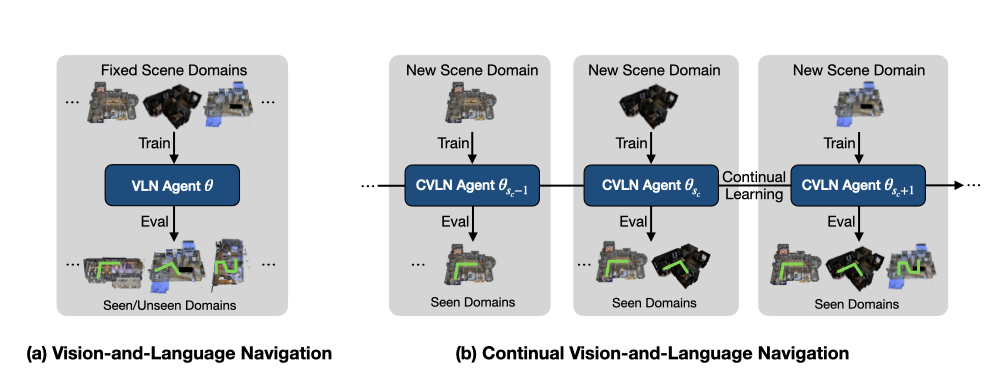} 
    \caption{Comparison between (a) VLN and (b) CVLN: VLN trains agents on fixed environments and evaluates them on unseen ones, while CVLN trains agents sequentially on new environments and evaluates them on both previously encountered and newly learned environments.}
    \label{fig:cvln_overview}
\end{figure*}

In this paper, we propose Continual Vision-and-Language Navigation (CVLN) paradigm, which trains and evaluates VLN agents with consideration of the CL ability. In CVLN, agents explore different \textit{scene domains}—each comprising various indoor scenes with multiple navigation episodes—sequentially and are evaluated on their ability to navigate through all of them. The CVLN agents need to retain their ability on previously learned scene domains while learning the current scene domain. Figure~\ref{fig:cvln_overview} illustrates an overview of CVLN. The CVLN supports two settings to handle different types of natural language instructions: Initial-instruction based CVLN (I-CVLN) and Dialogue based CVLN (D-CVLN). In I-CVLN, the instructions given to the agent at the start of navigation contain fine-grained information about the entire navigation path. In D-CVLN, the agent receives a target object and a conversation history between humans collaborating to locate it, and must reason navigation actions to reach the goal within environments.

We present two simple yet effective baseline methods for CVLN—Perplexity Replay (PerpR) and Episodic Self-Replay (ESR)—both based on a rehearsal mechanism, where the effective construction and utilization of replay memory are crucial. PerpR uses self-evaluated action perplexity (i.e. difficulty) to assess the importance of episodes, effectively selecting and organizing them in the replay memory based on their action perplexity. After learning a scene domain, the agent adds high-perplexity episodes to the replay memory and removes low-perplexity ones. Meanwhile ESR stores action logits predicted by the agent for each individual step within episodes in the replay memory. During subsequent training, the agent revisits these stored logits, allowing it to refine its learning by leveraging detailed episode step-level logits from past experiences. To validate the effectiveness of our methods, we conducted comparative experiments with existing CL methodologies. Furthermore, we conducted an ablation study, analyzed the effects of varying replay memory sizes, and examined the stability-plasticity trade-off.


\section{Related Work}

\noindent\textbf{Continual Learning.} Continual Learning (CL) focuses on training models to acquire new knowledge over time while retaining previously learned information. CL methods have generally been extensively studied for simple tasks~\cite{deng2009imagenet} in computer vision. This field includes several CL scenarios. Instance-Incremental Learning trains on batches of samples for the same task~\cite{lomonaco2017core50}. Domain-Incremental Learning (DIL) deals with tasks having the same labels but different input distributions, with no domain identity during inference~\cite{hsu2018re}. Task-Incremental and Class-Incremental Learning involve different label spaces, with varying use of domain identity~\cite{hsu2018re}. Task-Free Continual Learning handles separate label spaces without domain identity~\cite{aljundi2019task}, and Online Continual Learning processes sequential data streams in real-time~\cite{aljundi2019gradient}. CVLN fits within DIL, providing scene domain identities during training but not during testing, requiring generalization without explicit domain knowledge at evaluation. There are three main approaches for CL: Regularization~\cite{kirkpatrick2017overcoming}, which adds terms to prevent the loss of prior knowledge; Rehearsal~\cite{rebuffi2017icarl, chaudhry2018efficient, buzzega2020dark, boschini2022class}, which stores and reuses data from past tasks; and Architectural~\cite{rusu2016progressive, madotto2020continual}, which introduces task-specific parameters to the agent's architecture. This paper proposes two Rehearsal-based methods, PerpR and ESR, and evaluates their effectiveness through comparative experiments.

\noindent\textbf{Vision-and-Language Navigation.} In VLN, agents navigate realistic 3D indoor environments by following natural language instructions. Various datasets like R2R~\cite{anderson2018vln}, R4R~\cite{jain2019r4r}, and RxR~\cite{ku2020rxr} have been developed, focusing on navigation based on initial instructions that fully describe the path. Other tasks, such as CVDN~\cite{thomason2020vdn} and HANNA~\cite{nguyen2019help}, involve agents obtaining navigation information through dialogue with an oracle or other agents. Additionally, datasets like Touchdown~\cite{chen2019touchdown} and StreetLearn~\cite{mirowski2019streetlearn} have expanded VLN research to outdoor environments. IVLN~\cite{krantz2023iterative} present a paradigm where agents maintain memory across episodes to navigate a sequence of episodes. In the field of VLN, research focuses on improving agent architecture~\cite{hong2020recurrent, chen2021history, fried2018speaker, an2024etpnav}, enhancing visual-text alignment through auxiliary losses~\cite{ma2019self, wang2019reinforced, zhu2020vision}, using data augmentation~\cite{tan2019learning, wang2022less, li2022envedit, kamath2023new}, and extensive pre-training~\cite{hao2020towards,chen2021history, chen2022learning, an2023bevbert} to enhance generalization. In this paper, we adopted VLN-BERT~\cite{hong2020recurrent} and HAMT~\cite{chen2021history} as the backbone architectures. Existing VLN datasets and methods lack considerations for CL, highlighting the need for scalable strategies that enable agents to learn efficiently to diverse, changing environments without costly retraining models from scratch.

\section{Continual Vision-and-Lauguage Navigation}
\subsection{Formulation of VLN and CVLN}

In VLN, agents are trained on fixed data and evaluated on episodes from unseen scenes. Given a dataset \( D \) of episodes with navigation instructions \( I \) and ground-truth trajectories \( A^* = \{a_1^*, \ldots, a_N^*\} \), the agent \(\pi_\theta\) sequentially predicts actions based on visual observations \( V = \{v_1, \ldots, v_N\} \) and instructions \( I \). The objective minimizes the navigation imitation loss \(\ell\), typically cross-entropy:

\begin{equation}
    \mathcal{L}_D \triangleq \mathbb{E}_{(I,A^*) \sim D}[\ell(\pi_\theta(V, I), A^*)]
    \label{eq:VLN_CVLN}
\end{equation}

CVLN extends this by introducing continual learning across \( S \) scene domains. In each domain \( s \), episodes are drawn from distribution \( D_s \), and the agent must learn new domains without forgetting previous ones. The learning objective becomes:

\begin{equation}
    \underset{\theta}{\text{argmin}} \quad \sum_{s=1}^{s_c} \mathcal{L}_{D_s}, \quad \text{where} \ \mathcal{L}_{D_s} \triangleq \mathbb{E}_{(I,A^*) \sim D_s}[\ell(\pi_\theta(V, I), A^*)]
    \label{eq:CVLN}
\end{equation}

This continual learning setting better reflects real-world scenarios where agents encounter and must adapt to new environments over time.

\begin{figure*}[t]
\centering
\includegraphics[width=0.8\textwidth]{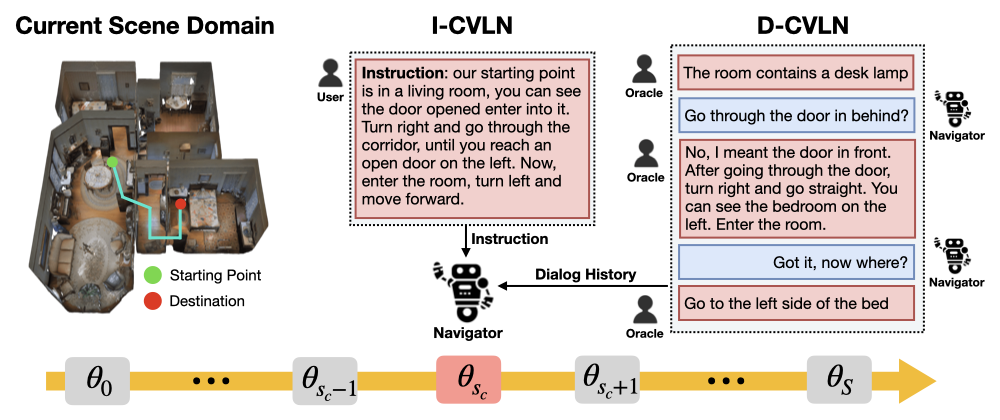} 
    \caption{Comparison between I-CVLN and D-CVLN. In I-CVLN, the agent is given an initial instruction containing all the information about the navigation path. Conversely, in D-CVLN, the agent obtains information about the navigation path through communication with an oracle.}
    \label{fig:fvnlclvsCVDNcl}
\end{figure*}

\noindent where an agent \(\pi\), with parameters \(\theta\), is optimized on one scene domain at a time in a sequential manner. \(s_c\) represents the current scene domain.

\subsection{I-CVLN and D-CVLN Benchmarks}

We design two CVLN benchmarks: Initial-Instruction based CVLN (I-CVLN) and Dialogue-based CVLN (D-CVLN), constructed from existing VLN datasets. In I-CVLN, agents receive an initial instruction describing the full navigation path. Using the R2R~\cite{anderson2018vln} and RxR~\cite{ku2020rxr} datasets, we construct the benchmark with English-only episodes. The training split from these datasets provides 15,700 episodes, while validation uses 1,563 unseen episodes from the same scenes (val-seen split). Episodes are grouped into 20 \textit{scene domains}, each containing over 600 training episodes. In D-CVLN, agents navigate based on dialogue history. This benchmark is built using the CVDN~\cite{thomason2020vdn} dataset, which includes dialogues between a follower and an oracle paired with corresponding trajectories. Due to limited episodes per scene, four scenes are grouped into each \textit{scene domain}, resulting in 11 domains. The training split includes 3,737 episodes, and validation uses 382 unseen episodes from the same domains. In both benchmarks, scene domains are learned sequentially, aligning with the continual learning setting.




\subsection{Evaluation of CVLN}

CVLN evaluation measures an agent’s ability to retain prior knowledge while adapting to new scene domains. We use the Average Metric, defined as:

\begin{equation}
  \frac{1}{S} \sum_{i=1}^{S} R_{S,i}
  \label{eq:AM}
\end{equation}

\noindent where \( R_{S,i} \) is the agent’s performance on the \( i \)th domain after learning the \( S \)th domain. For I-CVLN, \( R_{S,i} \) uses standard VLN metrics~\cite{anderson2018evaluation}: Success Rate (SR), Success weighted by inverse Path Length (SPL), and Navigation Error (NE). For D-CVLN, we use Goal Progress (GP) in meters from CVDN~\cite{thomason2020vdn}.

\section{Methods}

\begin{figure*}[t]
\centering
\includegraphics[width=0.9\textwidth]{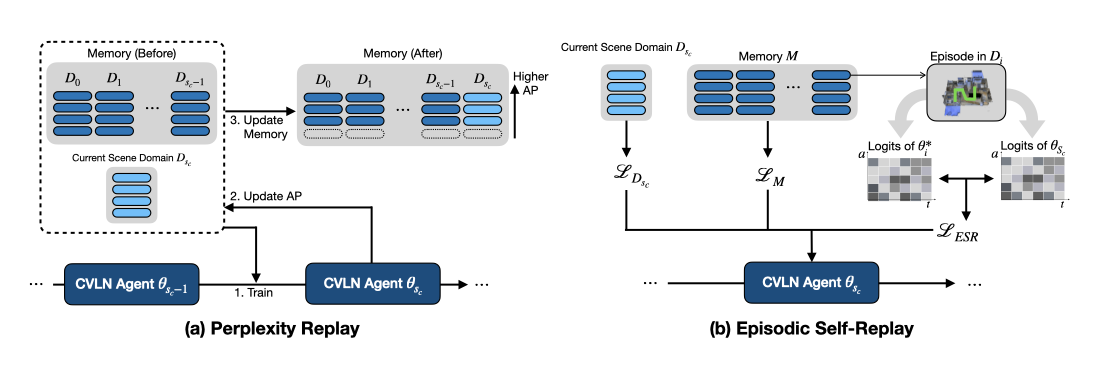}
    \caption{Overview of Perplexity Replay (PerpR) and Episodic Self-Replay (ESR) for CVLN: (a) PerpR prioritizes challenging episodes with high Action Perplexity (AP), and (b) ESR enables self-replay using past behaviors.}
    \label{fig:methods_overview}
\end{figure*}

Rehearsal-based CL approaches mitigate catastrophic forgetting by using a fixed-size replay memory \(M\) to store and revisit episodes from previous domains. After training each domain, \(M\) is updated by deleting an equal number of old episodes and adding new ones from the latest domain. Effective replay memory construction and usage are crucial for maintaining performance across domains. In this context, we propose Perplexity Replay (PerpR) and Episodic Self-Replay (ESR) as baseline methods for the CVLN agent. Figure~\ref{fig:methods_overview} provides an overview of PerpR and ESR.

\subsection{Perplexity Replay}

Random sampling~\cite{vitter1985random,riemer2018learning} is often used to construct replay memory for continual learning but overlooks sample importance for recovering past knowledge, leading to suboptimal memory composition. Existing methods~\cite{shim2021online,aljundi2019gradient} address this using gradient-based or optimizable approaches, but they are mainly designed for simple tasks like image classification and are unsuitable for sequential action prediction required in CVLN.




To address these limitations, we propose Perplexity Replay (PerpR), which builds replay memory by quantifying episode difficulty. PerpR introduces Action Perplexity (AP) to let the model self-evaluate episode difficulty without external metrics or annotations. Using AP, the model identifies challenging episodes with high uncertainty and prioritizes them in replay memory, focusing learning on harder experiences and improving overall performance.

Assuming episodes with high AP indicate insufficient agent learning, PerpR ensures that the replay memory is populated with experiences most beneficial for the agent's continued improvement. The replay memory update procedure involves selecting episodes with low AP for removal and adding those with high AP from previously trained scene domains. This self-directed mechanism allows the agent to concentrate on challenging episodes it has identified post-training, thereby improving its performance. The Action Perplexity is computed as:

\begin{equation}
  \text{AP}_{\theta}(\epsilon) = \exp\left({-\frac{1}{N}\sum_{i=1}^{N} \log P_{\theta}(a_i | v_i, I)}\right)
  \label{eq:AP}
\end{equation}

\noindent where \(N\) is the number of steps in an episode, \(v_i\) is the observed visual state at step \(i\), and \(P(a_i | v_i, I, \theta)\) is the probability of predicted action \(a_i\) of agent $\pi$ parameterized by $\theta$ based on \(v_i\), the instructions \(I\). 

We combine the replay memory with data from the current scene domain for the training of subsequent scene domains. The loss to be minimized is formulated as follows:

\begin{equation}
    \mathcal{L}_{PerpR} = \mathcal{L}_{D_{s_c}} + \mathcal{L}_{M}
\end{equation}
\begin{align*}
    \mathcal{L}_{D_{s_c}} &\triangleq \mathbb{E}_{(I,A^*) \sim D_{s_c}}[\ell(\pi_\theta(V, I), A^*)] \\
    \mathcal{L}_{M} &\triangleq \mathbb{E}_{(I,A^*) \sim M}[\ell(\pi_\theta(V, I), A^*)]
\end{align*}

\noindent where \(\mathcal{L}_{D_{s_c}} + \mathcal{L}_{M}\) represents the sum of the losses over the current scene domain \(D_{s_c}\) and the replay memory \(M\). The memory update procedure for PerpR is provided in the supplementary material (Section~4.1).

\subsection{Episodic Self-Replay}

To leverage replay memory effectively, prior works~\cite{rebuffi2017icarl, buzzega2020dark, boschini2022class, verwimp2022re} used logit distillation for non-sequential tasks like image classification. However, these methods are unsuitable for CVLN, which requires sequential action prediction. CVLN agents must handle temporally dependent observations and decision-making, while existing distillation methods focus only on single-step predictions, limiting their ability to learn complex action sequences.


To address this limitation, we extended these approaches for application to CVLN episodes. At each action step, we stored the action logits in a replay memory. When learning from episodes in the replay memory, the CVLN agent utilizes these stored logits to train itself to follow its past action distribution at each step. To reinforce consistency, we applied student forcing~\cite{sauder2020best} during training by using the stored logits as the next action for subsequent steps, rather than the predicted logits. This encourages adherence to prior action patterns and aligns the learning process with the sequential nature of CVLN.

We define the agent parameters immediately after learning the scene domain $s$ as $\theta_{s}^{*}$. A function $f$ outputs the logits for action candidates at each sequential step of navigation. Unlike other rehearsal-based methods, ESR stores the sequential action logits $Z \triangleq \{z_n\}_{n=1}^N \triangleq f_{\theta_{s}^{*}}(V, I)$, corresponding to each step in the sequence, when updating the replay memory $M$. ESR's replay memory uses random sampling~\cite{vitter1985random} to select episodes to add and delete.

The ESR loss function can be formulated as follows:

\begin{equation}
    \mathcal{L}_{ESR}\triangleq \mathbb{E}_{(I, Z) \sim M}\left[ \sum_{n=1}^{N} \|z_n - f_\theta(V_n, I_n)\|^2_F \right]
\end{equation}

ESR uses \(\mathcal{L}_{ESR}\) to allow agents to replay their behavior distribution during the learning process, which allows them to learn about new scene domains while effectively retaining previously learned knowledge. This helps mitigate catastrophic forgetting and improves the agent's ability to learn and adapt efficiently in a constantly changing environment. Also, ESR employs \(\mathcal{L}_{M}\), utilizing ground truth actions from replay memory. The final loss function of ESR is defined as follows:

\begin{equation}
  \mathcal{L}_{ESR\_total} = \mathcal{L}_{D_{s_c}} + \lambda_1 \mathcal{L}_{M} + \lambda_2 \mathcal{L}_{ESR}
  \label{eq:ESR}
\end{equation}

\noindent where \(\lambda_1\) and \(\lambda_2\) control the impact of \(\mathcal{L}_{M}\) and \(\mathcal{L}_{ESR}\) on the overall loss. Both \( \lambda_1 \) and \( \lambda_2 \) are set to 0.2. Please refer to supplementary material (Section~4.2) for more details.

\section{Experiments}

\subsection{Comparative Methods}

To establish the upper limit of our results, we provide the results of \textbf{Joint} agent trained across all scene domains simultaneously. Conversely, for the lower limit, we provide the result of \textbf{Vanilla} agent trained on scene domains sequentially without incorporating any CL methods. 

\noindent\textbf{L2 Regularization} reduces the difference between the current parameter \(\theta_{s}\) and the previous \(\theta_{s-1}\) by adding the regularization term \(\lambda \lVert \theta_s - \theta_{s-1}^* \rVert^2\) to the loss function.

\noindent\textbf{Random Replay (RandR)} uses Reservoir sampling~\cite{vitter1985random} to manage replay memory, integrating selected episodes with current training scene domain \(D_{s_c}\).

\noindent\textbf{A-GEM}~\cite{chaudhry2018efficient} limits gradient updates to minimize memory loss, improving computational complexity over GEM~\cite{lopez2017gradient}.

\noindent \textbf{AdapterCL} adds domain-specific adapters~\cite{houlsby2019parameter}; the adapter with the lowest action perplexity is selected during evaluation.

\subsection{Experiment Setups}

For I-CVLN, we use VLN-BERT~\cite{hong2020recurrent} with Oscar~\cite{li2020oscar} as the backbone and apply CL methods during training. The rehearsal-based method uses a fixed replay memory of 500 samples (about 3\% of the dataset), equally divided across 20 scene domains. AdapterCL employs 20 adapters matching these domains, expandable for more domains. Three curricula with different scene orders are generated using distinct seeds, and results are averaged; for non-comparative analyses, a single curriculum is used. For D-CVLN, we adopt a randomly initialized HAMT~\cite{chen2021history} backbone with CL methods. The replay memory is set to 100 samples (about 3\% of the dataset), and AdapterCL uses 11 fixed adapters. Three curricula covering 11 scene domains are configured, and results are averaged.

\begin{table*}[htbp]
\centering
\resizebox{0.7\textwidth}{!}{ 
\begin{tabular}{@{\extracolsep{1pt}}clcccc@{}}
\toprule
 & \multirow{2}{*}{Method} & \multicolumn{3}{c}{I-CVLN} & \multicolumn{1}{c}{D-CVLN} \\ 
\cmidrule{3-5} \cmidrule{6-6}
\ && AvgSPL$\uparrow$ & AvgSR$\uparrow$ & AvgNE$\downarrow$ & AvgGP$\uparrow$ \\ 
\midrule
1. & Vanilla     & $23.1 \pm 0.8$  & $25.6 \pm 0.6$  & $10.5 \pm 0.5$ & $5.5 \pm 0.4$ \\
2. & Joint       & $40.1 \pm 0.2$  & $42.6 \pm 0.1$  & $7.1 \pm 0.3$  & $8.1 \pm 0.1$ \\
\midrule
3. & L2          & $13.3 \pm 0.2$  & $14.6 \pm 0.2$  & $12.7 \pm 0.3$ & $4.3 \pm 0.4$ \\
4. & AdapterCL   & $7.8 \pm 0.3$   & $9.1 \pm 0.6$   & $13.9 \pm 0.8$ & $4.3 \pm 1.2$ \\
5. & AGEM        & $4.6 \pm 2.8$   & $4.5 \pm 2.2$   & $13.0 \pm 0.2$ & $2.7 \pm 0.9$ \\
6. & RandR       & $24.9 \pm 0.6$  & $28.0 \pm 0.7$  & $10.0 \pm 0.5$ & $5.7 \pm 0.3$ \\
7. & PerpR (Ours)& $\underline{26.1} \pm 1.6$ & $\underline{28.9} \pm 1.8$ & $\underline{9.8} \pm 0.2$ & $\textbf{6.1} \pm 0.4$ \\
8. & ESR (Ours)  & $\textbf{28.2} \pm 0.4$ & $\textbf{31.9} \pm 0.9$ & $\textbf{9.2} \pm 0.3$ & $\underline{5.8} \pm 0.5$ \\
\bottomrule
\end{tabular}
}
\caption{Comparative results of various methods on I-CVLN and D-CVLN. This table reports the mean and standard deviation values across three distinct learning curricula.}
\label{tab:main_comparison}
\end{table*}

\subsection{Comparative Results}
\label{subsec:main_result}

The results in Table~\ref{tab:main_comparison} provide a comparative analysis of various methods on I-CVLN and D-CVLN, highlighting the impact of continual learning (CL) strategies in overcoming catastrophic forgetting during sequential training. 

\noindent\textbf{Performance Bounds}: Vanilla and Joint define performance bounds, with Vanilla as the lower bound due to no forgetting mitigation, and Joint as the upper bound with full data access. Their large performance gap highlights the challenge of continual learning in CVLN, where new scene domains introduce unique challenges and increase forgetting risk.

\noindent\textbf{Performance of Traditional CL Approaches}: Standard CL methods such as L2, A-GEM, and AdapterCL fall short of even the Vanilla baseline, indicating that traditional CL strategies struggle with the complex, dynamic environments of I-CVLN and D-CVLN. This result demonstrates that while these methods can help manage forgetting in simpler tasks, they lack the robustness required for the sequential, environment-dependent learning in CVLN.

\noindent\textbf{Effectiveness of Rehearsal-Based Approach}: RandR achieves a noticeable improvement over other traditional methods, showcasing the importance of rehearsal for knowledge retention in sequential training. However, its performance still falls short of Joint, suggesting that further refinement in rehearsal strategies could yield more significant gains.

\noindent\textbf{Proposed Methods – PerpR and ESR}: Our methods, PerpR and ESR, build upon the strengths of RandR and outperform other CL methods across all metrics. These results indicate that our proposed approaches are effective in retaining previously learned knowledge and adapting to new scene domains. Notably, ESR excels on I-CVLN, while PerpR achieves the best results on D-CVLN, suggesting that the optimal method may vary depending on the specific evaluation setting. These findings highlight the adaptability and robustness of PerpR and ESR in handling the unique demands of CVLN tasks. 


\subsection{Analysis}
\textbf{Memory Size Analysis} The proposed rehearsal-based methods use replay memory, and their performance depends on its size. Table~\ref{tab:memorysize_comparison} shows that PerpR and ESR perform better with larger memory on both datasets, indicating that larger memory helps retain knowledge by storing more diverse samples.

\begin{table*}[t]
\centering
\resizebox{0.7\textwidth}{!}{ 
\begin{tabular}{@{}ccccccc@{}}
\toprule
\multirow{2}{*}{Method} & \multicolumn{4}{c}{I-CVLN} & \multicolumn{2}{c}{D-CVLN} \\ 
\cmidrule{2-5} \cmidrule{6-7}
& Memory Size & AvgSPL$\uparrow$ & AvgSR$\uparrow$ & AvgNE$\downarrow$ & Memory Size & AvgGP$\uparrow$ \\ 
\midrule
    & 200  & 23.8 & 26.3 & 10.2  & 50  & 4.9 \\
PerpR & 500  & 23.8 & 26.4 & 9.8   & 100 & \textbf{6.2} \\
    & 1000 & \textbf{28.7} & \textbf{31.1} & \textbf{8.7}  & 200 & 6.0 \\ 
\midrule
    & 200  & 25.8 & 29.2 & 9.2   & 50  & 5.3 \\
ESR & 500  & 27.7 & 31.1 & \textbf{8.8}   & 100 & 5.5 \\
    & 1000 & \textbf{29.8} & \textbf{34.4} & 9.1   & 200 & \textbf{5.9} \\ 
\bottomrule
\end{tabular}
}
\caption{Comparative analysis of the impact of replay memory size on performance for rehearsal-based methods. Larger memory sizes improve performance, indicating a positive correlation between buffer size and agent performance.}
\label{tab:memorysize_comparison}
\end{table*}

\begin{figure}[t]
\centering
\includegraphics[width=0.7\columnwidth]{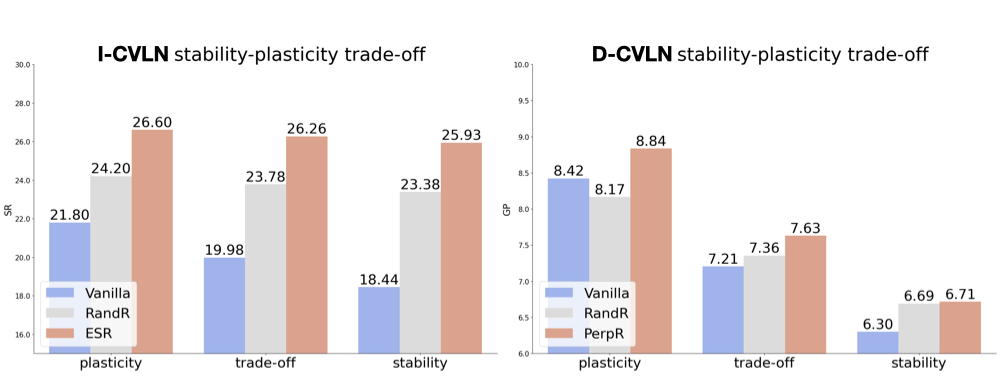} 
\caption{Stability-plasticity trade-off comparison in I-CVLN and D-CVLN: We calculate stability and plasticity for agents after learning 10 scene domains. }
\label{fig:SPT}
\end{figure}

\noindent\textbf{Stability-Plasticity Trade-off Analysis} In CVLN, agents require both memory stability to retain past knowledge and learning plasticity to acquire new information. These are often in trade-off, known as the stability-plasticity dilemma~\cite{parisi2019continual}. Stability (S) measures average performance on past domains after learning a new one, while Plasticity (P) measures initial performance on new domains. Catastrophic forgetting occurs when plasticity outweighs stability. We assess this trade-off~\cite{sarfraz2022synergy} using the harmonic mean of S and P (Figure~\ref{fig:SPT}). ESR achieves a better trade-off in I-CVLN than RandR and Vanilla, while PerpR shows comparable performance in D-CVLN.

\subsection{Ablation Study}




\textbf{Reversing PerpR memory update} Table~\ref{tab:perpr_reverse} shows the results of reversing the replay memory update in PerpR. PerpR is developed based on the assumption that episodes with higher action perplexity would be more beneficial for future learning. The results confirm the validity of PerpR's underlying assumption. PerpR's performance degrades when replay memory updates are reversed by storing low action perplexity episodes and deleting high ones.

\setlength{\tabcolsep}{1mm}
\begin{table}[t]
\centering
\begin{minipage}[t]{0.48\linewidth}
\centering
\resizebox{\linewidth}{!}{
\begin{tabular}{@{}lcccc@{}}
\toprule
\multirow{2}{*}{Ablation} & \multicolumn{3}{c}{I-CVLN} & \multicolumn{1}{c}{D-CVLN} \\ 
\cmidrule(lr){2-4} \cmidrule(l){5-5}
 & {SPL$\uparrow$} & {SR$\uparrow$} & {NE$\downarrow$} & {GP$\uparrow$} \\ 
\midrule
PerpR  & $\textbf{26.1} \pm 1.6$ & $\textbf{28.9} \pm 1.8$ & $\textbf{9.8} \pm 0.2$ &  $\textbf{6.1} \pm 0.4$ \\
PerpR-Rev.   & $24.6 \pm 1.1$ & $27.1 \pm 1.2$ & $10.2 \pm 0.3$ & $5.4 \pm 0.2$ \\
\bottomrule
\end{tabular}
}
\caption{Impact of memory update reversal on performance in PerpR.}
\label{tab:perpr_reverse}
\end{minipage}
\hfill
\begin{minipage}[t]{0.48\linewidth}
\centering
\resizebox{\linewidth}{!}{
\begin{tabular}{@{}lcccc@{}}
\toprule
\multirow{2}{*}{Ablation} & \multicolumn{3}{c}{I-CVLN} & \multicolumn{1}{c}{D-CVLN} \\ 
\cmidrule{2-4} \cmidrule{5-5}
 & {AvgSPL$\uparrow$} & {AvgSR$\uparrow$} & {AvgNE$\downarrow$} & {AvgGP$\uparrow$} \\ 
\midrule
$\mathcal{L}_{ESR\_total}$  & \textbf{27.7} & \textbf{31.1} & \textbf{8.8} &  \textbf{5.5} \\
- $\mathcal{L}_{M}$         & 26.0 & 27.9 & 9.7 & 5.4 \\ 
- $\mathcal{L}_{ESR}$       & 22.7 & 24.9 & 10.4 & 5.1 \\
\bottomrule
\end{tabular}
}
\caption{Ablation study on ESR for I-CVLN and D-CVLN. Effects of removing specific loss components.}
\label{tab:esr_ablation}
\end{minipage}
\end{table}


\noindent\textbf{ESR loss ablation} Table~\ref{tab:esr_ablation} shows the results of an ablation study on the ESR loss in Equation~\ref{eq:ESR}. The removal of each of the two losses shows decreased performance across all datasets. Notably, the removal of $\mathcal{L}_{ESR}$ resulted in a greater performance drop than the removal of $\mathcal{L}_{M}$. This confirms that leveraging the action distribution predictions from previous models helps mitigate the catastrophic forgetting problem in CVLN agents. Additionally, we observe a synergistic effect when using both types of losses obtained through replay memory.


\section{Conclusions and Future Work}

In this paper, we introduced the Continual Vision-and-Language Navigation paradigm to enable VLN agents to learn in changing environments without forgetting previous knowledge. We proposed two settings, I-CVLN and D-CVLN, to account for various forms of natural language instruction. Additionally, we present two simple yet effective methods—Perplexity Replay and Episodic Self-Replay—that use rehearsal mechanisms to mitigate catastrophic forgetting and are applicable to tasks like CVLN, which involve sequential decision-making. Our experiments showed that these methods outperform existing continual learning approaches, highlighting the importance of effective replay memory in CVLN. For future work, we plan to extend CVLN to continuous environments and design new settings and datasets specifically for VLN, beyond existing dataset-based benchmarks.


\noindent\textbf{Acknowledgements}
This work was partly supported by the Institute for Information \& Communications Technology Planning \& Evaluation (IITP) under grants RS-2021-II212068-AIHub (10\%), RS-2021-II211343-GSAI (10\%), RS-2022-II220951-LBA (15\%), and RS-2022-II220953-PICA (15\%); the National Research Foundation of Korea (NRF) under grants RS-2024-00353991-SPARC (15\%) and RS-2023-00274280-HEI (15\%); the Korea Evaluation Institute of Industrial Technology (KEIT) under grant RS-2024-00423940 (10\%); and the Gwangju Metropolitan City (Artificial Intelligence Industrial Convergence Cluster Development Project, 10\%) funded by the Korean government.

\bibliography{egbib}
\end{document}